\documentclass[11pt]{article}
\usepackage{color}
\usepackage{algorithm, algpseudocode}
\usepackage{mathrsfs}
\usepackage{dsfont}
\usepackage{lmodern}
\usepackage{array}
\usepackage{bbm}
\usepackage{amsmath}
\usepackage{amssymb}
\usepackage[mathscr]{eucal}
\usepackage{graphicx}
\usepackage{mathrsfs}
\usepackage{psfrag}
\usepackage{color}
\usepackage{here}
\usepackage[margin=1in]{geometry}
\usepackage{wasysym}
\usepackage{enumitem}
\usepackage{xcolor}
\usepackage{caption}
\usepackage{subcaption}
\usepackage{amsthm}
\usepackage{lipsum}
\usepackage[round]{natbib}
\graphicspath{{../media/}}

\definecolor{darkblue}{RGB}{0,0,140}
\usepackage[colorlinks=true,
            allcolors=darkblue]{hyperref}

\newtheorem{theorem} {Theorem}

\newtheorem{proposition} {Proposition}
\newtheorem{lemma} {Lemma}
\newtheorem{corollary} {Corollary}

\newtheorem{definition} {Definition}

\newtheorem{example} {Example}

\newcommand{\Exp}{\mathds{E}}

\newcommand{\Prob}{\mathds{P}}
\newcommand{\Real}{\mathds{R}}
\newcommand{\Nat}{\mathbb{N}}
\newcommand{\Ind}{\mathds{1}}

\newcommand{\so}{\succcurlyeq_{\rm so}}
\newcommand{\fsd}{\succcurlyeq_{\rm fsd}}
\newcommand{\ssd}{\succcurlyeq_{\rm ssd}}
\newcommand{\single}{\succcurlyeq_{\rm sc}}

\newcommand{\Fc}{\mathcal{F}}

\title{Gaussian-Dirichlet Posterior Dominance in Sequential Learning}

\author{
Ian Osband \\
Stanford University, DeepMind\\
\texttt{iosband@stanford.edu}
\and
Benjamin Van Roy \\
Stanford University\\
\texttt{bvr@stanford.edu}
}
\begin{document}
\maketitle

\begin{abstract}
We consider the problem of sequential learning from categorical observations bounded in $[0,1]$.
We establish an ordering between the Dirichlet posterior over categorical outcomes and a Gaussian posterior under observations with $N(0,1)$ noise.
We establish that, conditioned upon identical data with at least two observations, the posterior mean of the categorical distribution will always second-order stochastically dominate the posterior mean of the Gaussian distribution.
These results provide a useful tool for the analysis of sequential learning under categorical outcomes.
\end{abstract}

\section{Introduction}

For any $S \in \Nat$, fix any $v \in [0,1]^S$ and consider probabilities $P_1,\ldots,P_S$ associated with components of $v$.
Let the vector $P$ of probabilities itself be random and Dirichlet-distributed with parameters $\alpha \in \Real_{++}^S$.
Let $X | P$ be a vector of $n$ independent samples drawn from the associated categorical distribution over components of $v$.
Note that the  components of $X$ are conditionally independent, conditioned on $P$, but are not unconditionally independent.
Conditioned on $P$, the mean of each $X_n$ is $\overline{X} = P^\top v$.
Let $z \in [0,1]^n$ and $c_s = \sum_{i=1}^n \Ind \{z_i = v_s \}$ for each $s=1,..,S$.
Then, the distribution of $P$ conditioned on $X=z$ is Dirichlet with parameters $\alpha + c$.

Let $\overline{Y}$ be distributed ${\rm N}\left(\Exp[\overline{X}], (\alpha^\top {\bf 1})^{-1}\right)$.
Let $ Y | \overline{Y}$ be a vector of $N$ independent samples distributed according to ${\rm N}(\overline{Y}, 1)$.
The distribution of $\overline{Y}$ conditioned on $Y = z$ is ${\rm N}(\mu, \sigma^2)$, where
$$\mu = \frac{(\alpha+c)^\top v}{(\alpha+c)^\top {\bf 1}} \qquad \text{and} \qquad \sigma^2 = \frac{1}{(\alpha+c)^\top {\bf 1}}.$$

In this paper, we establish that
{\medmuskip=0mu \thinmuskip=0mu \thickmuskip=0mu $\overline{X} \mid (X=z) \ \ssd \ \overline{Y} \mid (Y=z)$},
where $\ssd$ denotes second-order stochastic dominance.
In other words, conditioned on identical outcomes, the posterior mean of the categorical distribution second-order-stochastically dominates the posterior mean of the Gaussian distribution.

This result extends earlier work relating variances of posterior means under Gaussian and Dirichlet models \citep{antoniak1974mixtures,kyung2009characterizing}.
Our result provides a dominance relation that applies to all moments.  Our interest in this result stems from its significance in the
area of \textit{reinforcement learning} \citep{Sutton1998}, where we have used it to establish a notion of stochastic optimism
achieved by particular reinforcement algorithms that generate randomized value functions to explore in an efficient
manner \citep{osband2014generalization,osband2016thesis}.
This paper presents the result and its proof in a form that will
be cited by our work on reinforcement learning and that will be accessible to researchers more broadly.

\section{Stochastic dominance}
\label{sec: stoch_dom}

In this section we will review several notions of partial orderings for real-valued random variables.
All random variables we define will be with respect to the probability space $(\Omega, \Fc, \Prob)$.

\begin{definition}[First order stochastic dominance (FSD)]
\label{def: fsd}
\hspace{0.0000001mm} \newline
Let $X$ and $Y$ be real-valued random variables.
We say that $X$ is (first order) stochastically dominant for $Y$ if for all $x \in \Real$,
\begin{equation}
  \Prob \left(X > x \right) \ge \Prob \left(Y > x \right).
\end{equation}
We write $X \fsd Y$ for this relationship.
\end{definition}

First order stochastic dominance defines a partial ordering between random variables but it also quite a blunt notion of dominance that will be insufficient for our purposes.
Consider $X \sim N(0, \sigma^2_X)$ and $Y \sim N(0, \sigma^2_Y)$ with $\sigma_X < \sigma_Y$.
These random variables cannot be related in terms of FSD.
However, in the context of gambling we might imagine that the return from $X$ is in some sense preferable to $Y$, since they have the same mean but $X$ is somehow less risky.
Our next definition formalizes this notion.

\begin{definition}[Second order stochastic dominance (SSD)]
\label{def: ssd}
\hspace{0.0000001mm} \newline
Let $X$ and $Y$ be real-valued random variables.
We say that $X$ is second order stochastically dominant for $Y$ if for all $u:\Real \rightarrow \Real$ concave and non-decreasing,
\begin{equation}
  \Exp[u(X)] \ge \Exp[u(Y)].
\end{equation}
We write $X \ssd Y$ for this relationship.
\end{definition}

\begin{proposition}[SSD equivalence]
\label{prop: ssd equiv}
\hspace{0.0001mm} \newline
Let $X$ and $Y$ be real-valued random variables with finite expectation.
The following are equivalent:
\begin{enumerate}[noitemsep, nolistsep]
    \item $X \ssd Y$

    \item For any $u:\Real \rightarrow \Real$ concave and increasing $\Exp[u(X)] \ge \Exp[u(Y)]$

    \item For any $\alpha \in \Real$, $\int_{-\infty}^{\alpha} \left\{ \Prob(Y \le s) - \Prob(X \le s) \right\} ds \ge 0$.


    \item $Y =_D X + A + W$ for $A \le 0$ and $\Exp\left[ W | X + A \right] = 0$ for all values $x + a$.

\end{enumerate}
\begin{proof}
This follows from a simple integration by parts \citep{hadar1969rules}.
\end{proof}
\end{proposition}

Second order stochastic dominance $X \ssd Y$ ensures that $\Exp[X] \ge \Exp[Y]$.
It also establishes that for \textit{any} convex loss $L$ that $X$ is less ``spread out'' than $Y$ in the sense $\Exp[L(X - \Exp[X])] \le \Exp[L(Y-\Exp[Y])]$.
Motivated by this equivalence, we introduce another related dominance condition.

\begin{definition}[Single crossing dominance (SCD)]
\label{def: single crossing}
\hspace{0.0001mm} \newline
Let $X$ and $Y$ be real-valued random variables with CDFs $F_X, F_Y$ and finite expectation.
We say that $X$ single-crossing dominates $Y$ if and only if $\Exp[X] \ge \Exp[Y]$ and there a crossing point $a \in \Real$ such that:
\begin{equation}
    F_Y(s) \ge F_X(s) \iff s \le a.
\end{equation}
We write $X \single Y$ for this relationship.
\end{definition}

Single crossing dominance is actually a stronger condition than SSD, as we show in Proposition \ref{prop: relate scd ssd}.
In general the reverse implication is not true, as we demonstrate in Example \ref{ex: ssd not sc}.

\begin{proposition}[SCD implies SSD]
\label{prop: relate scd ssd}
\hspace{0.0001mm} \newline
Let $X$ and $Y$ be real-valued random variables with finite expectation then
\begin{equation}
    X \single Y \implies X \ssd Y.
\end{equation}
\begin{proof}
Suppose $X \single Y$ with single crossing point $a$.
Let $I(\alpha) = \int_{-\infty}^{\alpha} \left\{ \Prob(Y \le s) - \Prob(X \le s) \right\} ds$.
By $X \single Y$ we know $I(\alpha) > 0$ for all $\alpha \le a$ and that $I(\alpha)$ is decreasing for all $\alpha \ge a$.
Now we consider the limit $\lim_{\alpha \rightarrow \infty} I(\alpha) = \Exp[X] - \Exp[Y] \ge 0$.
Hence $I(\alpha) \ge 0$ for all $\alpha \in \Real$, which shows that $X \so Y$ by Proposition \ref{prop: ssd equiv}.
\end{proof}
\end{proposition}

\begin{figure}[h!]
\centering
\includegraphics[width=0.85\linewidth]{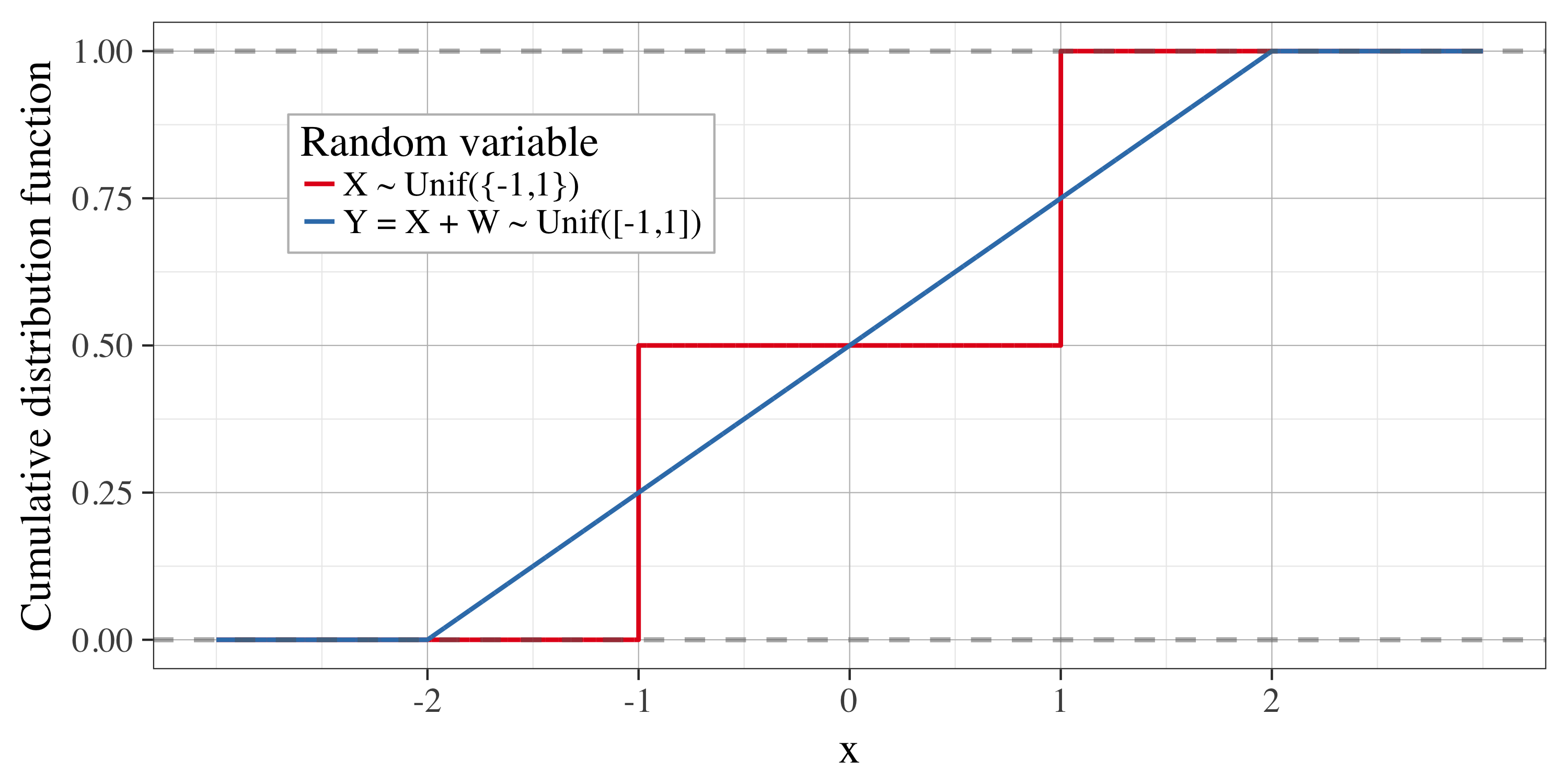}
\vspace{-3mm}
\caption{Second order stochastic dominance does not imply single crossing dominance}
\label{fig: double cross}
\end{figure}

\begin{example}[SSD does not imply SCD]
\label{ex: ssd not sc}
\hspace{0.0001mm} \newline
Consider $X \sim {\rm Unif}(\{-1, 1\})$ and let $Y = X + W$ where $W \sim {\rm Unif}([-1,1])$ and independent of $Y$.
By Proposition \ref{prop: ssd equiv} $X \ssd Y$, however $X$ is not single crossing dominant for $Y$.
\begin{proof}
We display the CDFs of these variables in Figure \ref{fig: double cross}, they are not single crossing.
In particular the ordering of $F_X, F_Y$ switches at least three points $x = -1, 0, 1$.
\end{proof}
\end{example}

\section{Gaussian-Dirichlet dominance}
\label{sec: gauss dir}

The main technical result in this paper comes in Theorem \ref{thm: dir_norm}, which we prove in Section \ref{sec: proof thm}.

\begin{theorem}[Gaussian vs Dirichlet dominance]
\label{thm: dir_norm}
\hspace{0.0001mm} \newline
Let $X = P^T v$ for $v \in [0,1]^S$ fixed and $P \sim {\rm Dirichlet}(\alpha)$ with $\alpha \in \Real^S_+$ and $\sum_{i=1}^S \alpha_i \ge 2$.
Let $Y \sim N(\mu, \sigma^2)$ with
$ \mu = \frac{\sum_{i=1}^S \alpha_i v_i}{\sum_{i=1}^S \alpha_i},
\ \sigma^2 = \left(\sum_{i=1}^S \alpha_i \right)^{-1}$, then $\Exp[X] = \Exp[Y]$ and $X \ssd Y$.
\end{theorem}

At first glance, Theorem \ref{thm: dir_norm} may seem quite arcane, it provides an ordering between two paired families of Gaussian and Dirichlet distributions in terms of SSD.
The reason this result is so useful is that, given matched prior distributions, the resultant posteriors for the Gaussian and Dirichlet models will remain ordered in this way \textit{for any} observation data.
The condition $\sum_{i=1}^S \alpha_i \ge 2$ is technical but does not pose significant difficulties so long as at the posterior is updated with at least two observations.
We present this result as Corollary \ref{cor: dir_norm}.

\begin{corollary}[Gaussian vs Dirichlet posterior ordering]
\label{cor: dir_norm}
\hspace{0.0001mm} \newline
Let $X = P^T v$ for $v \in [0,1]^S$ fixed and $P \sim {\rm Dirichlet}(\alpha)$ with $\alpha \in \Real^S_+$.
Let $Y \sim N(\mu, \sigma^2)$ with
$ \mu = \frac{\sum_{i=1}^S \alpha_i v_i}{\sum_{i=1}^S \alpha_i},
\ \sigma^2 = \left(\sum_{i=1}^S \alpha_i \right)^{-1}$.
Let $D$ be the data from $n$ i.i.d. samples from the categorical distribution $P$ and values $v$.
Let $\tilde{X}$ be the posterior distribution for $X \mid D$ and $\tilde{Y}$ be the posterior distribution for $Y \mid D$ but updating according to a mis-specified likelihood as if the observations were $\sim N(\mu, 1)$.
Then, for all datasets $D$ such that $n + \sum_{i=1}^S \alpha_i \ge 2$ we can guarantee that $\tilde{X} \ssd \tilde{Y}$.

\begin{proof}
This result is a consequence of Theorem \ref{thm: dir_norm} together with algebraic relations for the conjugate updates of $\tilde{X}$ and $\tilde{Y}$ given any data $D$.
We write $\vec{n} \in \Nat^S$ for the number of observations from each category $v$ in the dataset $D$ with $\sum_{s=1}^S \vec{n}_s = n$.
Then we can write the posterior distribution $\tilde{X} = X \mid D = P^T v$ for $P \sim {\rm Dirichlet}(\tilde{\alpha})$
and $\tilde{\alpha} := \alpha + \vec{n}$.

In a a similar way we can compute the posterior distribution of $\tilde{Y} = Y \mid D$ where we update with an misspecified likelihood as if the data were $\sim N(\mu, 1)$.
Once again we can use a conjugate form for the update $\tilde{Y} \sim N(\mu, \sigma^2)$ explicitly,
$$\mu = \frac{\sum_{s=1}^S \vec{n}_s v_s + \frac{\sum_{s=1}^S \alpha_s v_s}{\sum_{s=1}^S \alpha_s}\sum_{s=1}^S \alpha_s}{\sum_{s=1}^S \vec{n}_s + \sum_{s=1}^S \alpha_s}
= \frac{\sum_{s=1}^S \tilde{\alpha}_s v_s}{\sum_{s=1}^S \tilde{\alpha}_s},$$
$$\sigma^2 = \frac{1}{\sum_{s=1}^S \vec{n}_s + \sum_{s=1}^S \alpha_s}
= \frac{1}{\sum_{s=1}^S \tilde{\alpha}_s}.$$
We conclude by application of Theorem \ref{thm: dir_norm} on the updated posterior parameters $\tilde{\alpha}$.
\end{proof}
\end{corollary}




\section{Proof of Theorem \ref{thm: dir_norm}}
\label{sec: proof thm}
The complete proof of Theorem \ref{thm: dir_norm} is long but the essential argument is simple.
We outline the main arguments below and fill in the details in Sections \ref{app: beta dir} and \ref{app: gauss beta}.
First, we consider an auxilliary random variable $\tilde{X} \sim {\rm Beta}(\tilde{\alpha}, \tilde{\beta})$ with $\tilde{\alpha} = \sum_{i=1}^S \alpha_i v_i$ and $\tilde{\beta} = \sum_{i=1}^S \alpha_i - \tilde{\alpha}$.
In Lemma \ref{lem: Dir Beta} we show that $X \ssd \tilde{X}$.
Next, we show that this auxilliary beta $\tilde{X}$ is single crossing dominant for the approximating Gaussian posterior, $\tilde{X} \single Y$.
Therefore, by Proposition \ref{prop: relate scd ssd} $X \ssd Y$.

The main difficulty in this proof comes in establishing $\tilde{X} \single Y$.
To do this we use a laborious calculus argument together with repeated applications of the mean value theorem.
Our proof requires separate upper and lower bounds for different regions of $\tilde{\alpha}$ and $\tilde{\beta}$, but no real insight beyond that.
We believe that there should be a much more enlightened and elegant method to obtain these results.

\subsection{Beta vs Dirichlet}
\label{app: beta dir}

We begin our proof of Theorem \ref{thm: dir_norm} with an intermediate comparison of the Dirichlet distribution to a matched Beta posterior.
We first state a more basic result that we will use on Gamma distributions.
\begin{lemma}[Conditioning the sum of Gamma random variables]
\label{le:gamma}
\hspace{0.000001mm} \newline
Let $\gamma_1 \sim {\rm Gamma}(k_1, \theta)$ and $\gamma_2 \sim {\rm Gamma}(k_2, \theta)$ be independent random variables.
Then the conditional expectations
$\Exp[\gamma_1 | \gamma_1 + \gamma_2] = \frac{k_1}{k_1 + k_2} (\gamma_1+\gamma_2)$
and
$\Exp[\gamma_2 | \gamma_1 + \gamma_2] = \frac{k_2}{k_1 + k_2} (\gamma_1+\gamma_2).$
\end{lemma}

{
\begin{lemma}[Beta vs Dirichlet dominance]
\label{lem: Dir Beta}
\hspace{0.00000001mm} \newline
Let $X = P^\top v$ for the random variable $P \sim {\rm Dirichlet}(\alpha)$ and constants $v \in \Real^S$ and $\alpha \in \Real_+^S$.
Without loss of generality, assume $v_1 \leq v_2 \leq \cdots \leq v_S$.
Let $\tilde{\alpha} = \sum_{i=1}^S \alpha_i (v_i - v_1) / (v_S-v_1)$ and $\tilde{\beta} = \sum_{i=1}^S \alpha_i (v_S - v_i) / (v_S-v_1)$.
Then, there exists a random variable $\tilde{p} \sim {\rm Beta}(\tilde{\alpha},\tilde{\beta})$ such that,
for $\tilde{X} = \tilde{p} v_d + (1-\tilde{p}) v_1$, $\Exp[\tilde{X} | X] = \Exp[X]$ and so $X \ssd \tilde{X}$.
\end{lemma}
}


\begin{proof}
Let $\gamma_i = \text{Gamma}(\alpha, 1)$, with $\gamma_1,\ldots,\gamma_S$ independent, and let $\overline{\gamma} = \sum_{i=1}^S \gamma_i$, so that
$P \equiv_D \gamma / \overline{\gamma}.$
Let $\alpha_i^0 = \alpha_i (v_i - v_1) / (v_S-v_1)$ and $\alpha_i^1 = \alpha_i (v_S - v_i) / (v_S-v_1)$ so that $\alpha = \alpha^0 + \alpha^1.$
Define independent random variables
$\gamma^0 \sim \text{Gamma}(\alpha_i^0, 1)$ and $\gamma^1 \sim \text{Gamma}(\alpha_i^1, 1)$ so that
$\gamma \equiv_D \gamma^0 + \gamma^1.$

Take $\gamma^0$ and $\gamma^1$ to be independent, and couple these variables with $\gamma$ so that
$\gamma = \gamma^0 + \gamma^1.$
Note that $\tilde{\beta} = \sum_{i=1}^S \alpha^0_i$ and  $\tilde{\alpha} = \sum_{i=1}^S \alpha^1_i$.
Let $\overline{\gamma}^0 = \sum_{i=1}^S \gamma^0_i$ and $\overline{\gamma}^1 = \sum_{i=1}^S \gamma^1_i$, so that
$1-\tilde{p} \equiv_D \overline{\gamma}^0 / \overline{\gamma}$ and
$\tilde{p} \equiv_D \overline{\gamma}^1 / \overline{\gamma}.$
Couple these variables so that
$1-\tilde{p} = \overline{\gamma}^0 / \overline{\gamma}$ and
$ \tilde{p} = \overline{\gamma}^1 / \overline{\gamma}.$
We can now say
\begin{eqnarray*}
\Exp[\tilde{X} | X]
&=& \Exp[(1- \tilde{p}) v_1 + \tilde{p} v_d | X]
= \Exp\left[\frac{v_1 \overline{\gamma}^0}{\overline{\gamma}} + \frac{v_S \overline{\gamma}^1}{\overline{\gamma}} \Big| X\right] \\
&=& \Exp\left[\Exp\left[\frac{v_1 \overline{\gamma}^0 + v_S \overline{\gamma}^1}{\overline{\gamma}} \Big| \gamma, X\right] \Big| X \right]
= \Exp\left[\frac{v_1 \Exp[\overline{\gamma}^0 | \gamma] + v_S \Exp[\overline{\gamma}^1 | \gamma]}{\overline{\gamma}} \Big| X \right] \\
&=& \Exp\left[\frac{v_1 \sum_{i=1}^S \Exp[\gamma^0_i | \gamma_i] + v_S \sum_{i=1}^Sxp[\gamma^1_i | \gamma_i]}{\overline{\gamma}} \Big| X \right] \\
&\stackrel{\text{(a)}}{=}& \Exp\left[\frac{v_1 \sum_{i=1}^S \gamma_i \alpha_i^0 / \alpha_i + v_S \sum_{i=1}^S \gamma_i \alpha_i^1/\alpha_i}{\overline{\gamma}} \Big| X \right] \\
&=& \Exp\left[\frac{v_1 \sum_{i=1}^S \gamma_i (v_i - v_1) + v_S \sum_{i=1}^S \gamma_i (v_S - v_i)}{\overline{\gamma} (v_S - v_1)} \Big| X \right] \\
&=& \Exp\left[\frac{\sum_{i=1}^S \gamma_i  v_i}{\overline{\gamma}} \Big| X \right]
= \Exp\left[\sum_{i=1}^S p_i  v_i \Big| X \right]
= X,
\end{eqnarray*}
where (a) follows from Lemma \ref{le:gamma}.
Therefore, $\tilde{X}$ is a mean-preserving spread of $X$ and so by Proposition \ref{prop: ssd equiv}, $X \ssd \tilde{X}$.
\end{proof}

\subsection{Gaussian vs Beta}
\label{app: gauss beta}
We complete the proof of Theorem \ref{thm: dir_norm} by showing that this auxilliary Beta random variable $\tilde{X}$ defined in Lemma \ref{lem: Dir Beta} is second order stochastic dominant for the Gaussian posterior $Y$.

\begin{lemma}[Gaussian vs Beta dominance]
\label{lem: Gauss Beta}
\hspace{0.0000001mm} \newline
Let $\tilde{X} \sim {\rm Beta}(\alpha, \beta)$ for any $\alpha>0, \beta>0$ and $Y \sim N \left(\mu = \frac{\alpha}{\alpha + \beta}, \sigma^2 = \frac{1}{\alpha + \beta} \right)$.
Then, $\tilde{X} \single Y$ (and by Proposition \ref{prop: relate scd ssd} this implies $\tilde{X} \ssd Y$) whenever $\alpha + \beta \ge 2$.
\end{lemma}


We want to prove that the CDFs cross at most once on $(0,1)$.
By the mean value theorem \citep{rudin1964principles}, it is sufficient to prove that the PDFs cross at most twice on the same interval.
We lament that the proof as it stands is so laborious, but our attempts at a more elegant solution has so far been unsuccessful.
The remainder of this appendix is devoted to proving this ``double-crossing'' property via manipulation of the PDFs for different values of $\alpha, \beta$.

We write $f_N$ for the density of the Normal $Y$ and $f_B$ for the density of the Beta $\tilde{X}$ respectively.
We know that at the boundary $f_N(0-) > f_B(0-)$ and $f_N(1+) > f_B(1+)$ where the $\pm$ represents the left and right limits respectively.
As these densities are positive over the interval, we can consider the log PDFs
$$ l_B(x) = (\alpha - 1) \log(x) + (\beta - 1) \log(1-x) + K_B $$
$$ l_N(x) = -\frac{1}{2} (\alpha + \beta) \left(x - \frac{\alpha}{\alpha+\beta}\right)^2 + K_N.$$
The function $\log(x)$ is injective and increasing; if we can show that $l_N(x) - l_B(x) = 0$ has at most two solutions on the interval we will be done.

Instead we will attempt to prove an even stronger condition, that $l'_N(x) - l'_B(x) = 0$ has at most one solution in the interval.
This sufficient condition may be easier to deal with since we can ignore the distributional normalizing constants.
\begin{equation*}
   l'_B(x) = \frac{\alpha - 1}{x} - \frac{\beta - 1}{1 - x} \ , \ \ \
   l'_N(x) = \alpha - (\alpha + \beta) x
\end{equation*}

Finally we consider an even stronger condition, if $l''_N(x) - l''_B(x) = 0$ has no solution then $l'_B(x) - l'_N(x)$ must be monotone over the region and so it can have at most one root.
\begin{equation*}
    l''_B(x) = - \frac{\alpha - 1}{x^2} - \frac{\beta - 1}{(1-x)^2} \ , \ \ \
    l''_N(x) = - (\alpha + \beta)
\end{equation*}
With these definitions now let us define:
\begin{equation}
\label{eq: h_def}
    h(x) := l''_N(x) - l''_B(x) = \frac{\alpha - 1}{x^2} + \frac{\beta - 1}{(1-x)^2}
        - (\alpha + \beta)
\end{equation}
Our goal now is to show that $h(x) = 0$ does not have any solutions for $x \in [0,1]$.
Once again, we will look at the derivatives and analyze them for different values of $\alpha, \beta > 0$.
$$ h'(x) = -2 \left( \frac{\alpha - 1}{x^3} - \frac{\beta - 1}{(1-x)^3} \right) $$
$$ h''(x) = 6 \left( \frac{\alpha - 1}{x^4} + \frac{\beta - 1}{(1-x)^4} \right) $$


Our proof will proceed by considering specific ranges for the values of $\alpha, \beta > 0$ and use different calculus arguments for each of these regions.
By symmetry in the problem, we only need to prove the result for $\alpha > \beta$.
Within this section of possible parameter values we will need to subdivide the quadrant into three proof regions.
$R1 := \{ \alpha > 1 \ge \beta \ge 0 \}$, $R2 := \{ \alpha > 1, \beta > 1, (\alpha - 1)(\beta - 1) \ge 1/9\}$ and $R3 := \{ \alpha > 1, \beta > 1, (\alpha - 1)(\beta - 1) < 1/9\}$.
These regions completely cover all $\alpha + \beta \ge 2$ and hence suffice to complete the proof of Lemma \ref{lem: Gauss Beta}.

\begin{figure}[h!]
  \centering
  \includegraphics[width=0.8\linewidth]{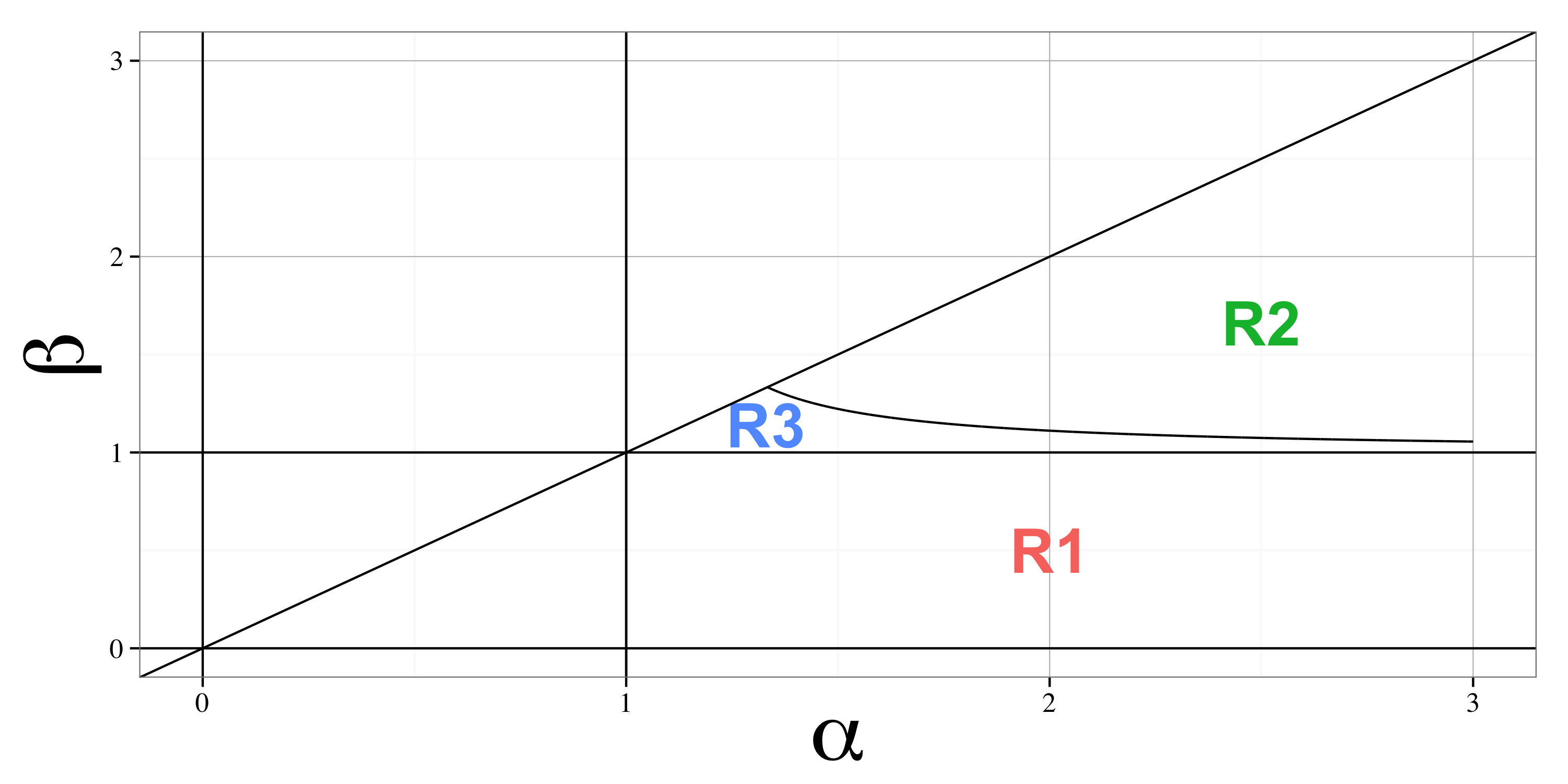}
  \vspace{-3mm}
  \captionsetup{justification=centering}
  \caption{Parameter regions for proof. The special case $\alpha = \beta = 1$ can be verified individually, in this case the PDFs do not intersect at any point.}
  \label{fig: proof regions}
\end{figure}

\subsubsection{Region $R1 = \{ \alpha \ge 1 \ge \beta \ge 0 \}$}
\label{sec: r1}
In this region we will show that $g(x) = l'_N(x) - l'_B(x)$ has no solutions.
We write $A = \alpha - 1 > 0$ and $B = \beta - 1 \le 0$ as before.
$$ g(x) = \alpha - (\alpha + \beta)x + \frac{\beta - 1}{1-x} - \frac{\alpha - 1}{x} $$
$$ g'(x) = h(x) = \frac{A}{x^2} + \frac{B}{(1-x)^2} - (\alpha + \beta) $$
$$ g''(x) = h'(x) = -2 \left( \frac{A}{x^3} - \frac{B}{(1-x)^3} \right) $$
We note that $g''(x) \le 0$ and so $g(x)$ is a concave function.
If we can show that the maximum of $g$ lies below $0$ then we know that there can be no roots.
We now attempt to solve $g'(x) = 0$:
\begin{eqnarray*}
    g'(x) = \frac{A}{x^2} + \frac{B}{(1-x)^2} = 0 \
    \implies \ - A / B = \left( \frac{x}{1-x} \right)^2 \
    \implies \ x = \frac{K}{1+K} \in (0,1),
\end{eqnarray*}
where here we write $K = \sqrt{-A/B} > 0$.
We ignore the case $B=0$ as a trivial special case.
We write $C = -B \ge 0$ and evaluate the function $g$ at its minimum $x_K = \frac{K}{1+K}$.
\begin{eqnarray*}
    g(x_K) &=& (A+1) - (A+B+2)\frac{K}{1+K} + B(1+K) - A\frac{1+K}{K} \\
    &=& - AK^2 - AK - A + BK^3 + BK^2 + BK - K^2 + K \\
    &=& - AK^2 - AK - A - CK^3 - CK^2 - CK - K^2 +K \\
    &=& - A (A/C) - A (A/C)^{1/2} - A - C (A/C)^{3/2} - C (A/C) - C(A/C)^{1/2} - A/C + (A/C)^{1/2} \\
    &=& -A^2 C^{-1} - A^{3/2}C^{-1/2} - A - A^{3/2}C^{-1/2} - A - A^{1/2}C^{1/2} - AC^{-1} + A^{1/2}C^{1/2} \\
    &=& -A^2 C^{-1} - 2 A^{3/2}C^{-1/2} - 2 A - AC^{-1} \le 0
\end{eqnarray*}
Therefore the Lemma holds for all $\alpha, \beta \in R1$

\subsubsection{Region $R2 = \{ \alpha > 1, \beta > 1, (\alpha - 1)(\beta - 1) \ge 1/9\}$}
\label{sec: r2}
In the case of $\alpha, \beta > 1$ we know that $h(x)$ is a convex function on $(0,1)$.
If we solve $h'(x^*) = 0$ and $h(x^*) > 0$ then we prove our statement.
We will write $A = \alpha - 1, B = \beta - 1$ for convenience.

First we solve $h'(x) = 0$ in terms of $K = \left( A / B \right)^{1/3} > 0$,
\begin{eqnarray*}
    h'(x) = \frac{A}{x^3} - \frac{B}{(1-x)^3} = 0 \
    \implies \ A / B = \left( \frac{x}{1-x} \right)^3 \
    \implies \ x = \frac{K}{1+K} \in (0,1).
\end{eqnarray*}
We can now evaluate the function $h$ at its minimum $x_K = \frac{K}{1+K}$.
\begin{eqnarray*}
    h(x_K) &=& A \frac{(K+1)^2}{K^2} + B(K+1)^2 - (A + B + 2) \\
    &=& A (2 / K + 1 / K^2) + B ( K^2 + 2K ) - 2 \\
    &=& 3 (A^{2/3}B^{1/3} + A^{1/3}B^{2/3}) - 2.
\end{eqnarray*}
As long as $h(x_K) > 0$ we have shown that the CDFs are single crossing.
We note that for all $\alpha, \beta \in R2$
$$ A,B \ge 1/3 \implies AB \ge 1/9 \implies (A^{2/3}B^{1/3} + A^{1/3}B^{2/3}) \ge 2/3 .$$
This completes the proof for $R2$.

\subsubsection{Region $R3 = \{ \alpha > 1, \beta > 1, (\alpha - 1)(\beta - 1) < 1/9\}$}
\label{sec: r3}

Our argument for this final region is no different than before, although it is slightly more involved.
The key additional difficulty is that it in this region is not enough to only look at the derivatives of the log likelihoods; we need to use some bound on the normalizing constants to get our bounds.

In $R3$, we know that $\beta \in (1,\frac{4}{3})$ so we will make use of an upper bound to the normalizing constant of the Beta distribution, the Beta function.
\begin{eqnarray}
    B(\alpha, \beta) &=& \int_{x=0}^1 x^{\alpha - 1}(1-x)^{\beta-1} dx
    \le \int_{x=0}^1 x^{\alpha-1} dx = \frac{1}{\alpha}
\end{eqnarray}
The intuition is that, because in $R3$ the value of $\beta-1$ is relatively small, this approximation will not be too bad.
Therefore, we can explicitly bound the log likelihood of the Beta distribution:
$$ l_B(x) \ge \tilde{l}_B(x) := (\alpha - 1) \log(x) + (\beta - 1) \log(1-x) + \log(\alpha) $$

We now repeat a familiar argument based upon explicit calculus.
We want to find two points $x_1 < x_2$ for which $h(x_i) = l''_N(x) - l''_B(x) > 0$.
Since $\alpha, \beta > 1$ we know that $h$ is convex and so for all $x \notin [x_1, x_2]$ then $h > 0$.
We define the gap of the Beta over the maximum of the normal log likelihood,
\begin{equation}
    {\rm Gap: } \ l_B(x_i) - l_N(x_i) \ge f(x_i) := \tilde{l}_B(x_i) - \max_x l_N(x) > 0.
\end{equation}
If we can show the gap is positive then it must mean there are no crossings over the region $[x_1, x_2]$.
This is because $\tilde{l}_B$ is concave and therefore totally above the maximum of $l_N$ over the whole region $[x_1, x_2]$.

Consider any $x \in [0,x_1)$; we know from the ordering of the tails of the CDF that if there is more than one root in this segment then there must be at least three crossings.
If there are three crossings, then the second derivative of their difference $h$ must have at least one root on this region.
However we know that $h$ is convex, so if we can show that $h(x_i) > 0$ this cannot be possible.
We use a similar argument for $x \in (x_2, 1]$ and complete this proof via laborious calculus.

We remind the reader of the definition in \eqref{eq: h_def},
$ h(x) := l''_N(x) - l''_B(x) = \frac{\alpha - 1}{x^2} + \frac{\beta - 1}{(1-x)^2}
        - (\alpha + \beta) $.
For ease of notation we will write $A = \alpha - 1, B = \beta - 1$.
We note that:
$$h(x) > h_1(x) = \frac{A}{x^2} - (A+B+2)$$
$$h(x) > h_2(x) = \frac{B}{(1-x)^2} - (A+B+2)$$
and we solve for $h_1(x_1) = 0, h_2(x_2)=0$.
This means that
$$x_1 = \sqrt{\frac{A}{A+B+2}} \ , \ x_2 = 1 - \sqrt{\frac{B}{A+B+2}}$$
and clearly $h(x_1) > 0, h(x_2) > 0$.
Now, if we can show that, for all possible values of $A,B$ in this region $f(x_i) = l_B(x_i) - \max_x l_N(x) > 0$, our proof will be complete.

To make the dependence on $A, B$ more clear we write $f(x_i) = f_i(A,B)$ below
{\medmuskip=1mu
\thinmuskip=1mu
\thickmuskip=1mu
{\small
$$ f_1(A, B) = \log(1 + A) + A \log\left( \sqrt{\frac{A}{A + B + 2}} \right) + B \log\left(1 - \sqrt{\frac{A}{A + B + 2}} \right) + \frac{1}{2} \log(2 \pi) - \frac{1}{2}\log(A + B + 2), $$
$$ f_2(A, B) = \log(1 + A) + A \log\left( 1 - \sqrt{\frac{B}{A + B + 2}} \right) + B \log\left(\sqrt{\frac{B}{A + B + 2}} \right) + \frac{1}{2} \log(2 \pi) - \frac{1}{2}\log(A + B + 2). $$
}
We will demonstrate that $\frac{\partial f_i}{\partial B} \le 0$ for all of the values in our region $A > B > 0$.
{\small
\begin{eqnarray*}
\frac{\partial f_1}{\partial B} &=& -\frac{A}{2(A + B +2)}
    + \log\left(1-\sqrt{\frac{A}{A + B + 2}}\right)
    + \frac{B \sqrt{A}}{2(A+B+2)^{3/2}\left(1-\sqrt{\frac{A}{A + B + 2}}\right)}
    - \frac{1}{2(A+B+2)} \\
    &=& \frac{1}{2(A+B+2)} \left(\frac{B \sqrt{A}}{\sqrt{A+ B +2}\left(1-\sqrt{\frac{A}{A + B + 2}}\right)} - A - 1 \right)
        + \log\left(1-\sqrt{\frac{A}{A + B + 2}}\right) \\
    &=& \frac{1}{2(A+B+2)} \left( \frac{B \sqrt{A}}{\sqrt{A + B +2} -\sqrt{A}} - A - 1 \right)
        + \log\left(1-\sqrt{\frac{A}{A + B + 2}}\right) \\
    &\le& \frac{1}{2(A+B+2)} \left( \frac{\sqrt{B} / 3}{\sqrt{A + B +2} -\sqrt{A}} - A - 1 \right) -\sqrt{\frac{A}{A + B + 2}} \\
    &\le& \frac{1}{2(A+B+2)} \left( \frac{1}{3} \sqrt{\frac{B}{B+2}} - A - 1 \right) -\sqrt{\frac{A}{A + B + 2}} \\
    &\le& - \frac{A}{2(A+B+2)} - \sqrt{\frac{A}{A + B + 2}} \ \ \le \  0.
\end{eqnarray*}}
Similarly,
{\small
\begin{eqnarray*}
\frac{\partial f_2}{\partial B} &=&
    - A \left(\frac{\sqrt{\frac{B}{A+B+2}}}{2B} + \frac{1}{2(A+B+2)} \right)
    + \log\left(\sqrt{\frac{B}{A + B + 2}} \right)
    + B \left( \frac{A + 2}{2B(A+B+2)}\right)
    - \frac{1}{2(A + B +2)} \\
    &=& \frac{1}{2(A+B+2)} \left(A+ 2 - A - 1 - A\sqrt{\frac{A+B+2}{B}} \right)
        + \log\left(\sqrt{\frac{B}{A + B +2}} \right) \\
    &=& \frac{1}{2(A+B+2)} \left(1 - A\sqrt{\frac{A+B+2}{B}} \right)
        + \frac{1}{2}\log\left(\frac{B}{A + B +2} \right).
\end{eqnarray*}}
Therefore, for any $A \ge 0$ this means that $\frac{\partial^2 f_2}{\partial A \partial B} < 0$.
Therefore this expression $\frac{\partial f_2}{\partial B}$ is maximized over $A$ for $A=0$.
We can evaluate this expression explicitly:
\begin{eqnarray*}
\frac{\partial f_2}{\partial B} \big\vert_{A=0}
    \ = \ \frac{1}{2(B+2)} + \frac{1}{2}\log\left(\frac{B}{B +2} \right)
    \ \le \ \frac{1}{2} \left(\frac{1}{B+2} + \frac{B}{B+2} - 1 \right)
    \ \le \ 0.
\end{eqnarray*}

This provides a monotonicity result which states that both $f_1, f_2$ are minimized at at the largest possible $B = \frac{1}{9A}$ for any given $A$ over our region.
We will now write $g_i(A) := f_i(A, \frac{1}{9A})$.
If we can show that $g_i(A) \ge 0$ for all $A \ge \frac{1}{3}$ and $i=1,2$ we will be done with our proof.
We will perform a similar argument to show that $g_i$ is monotone increasing for all $A \ge \frac{1}{3}$.
{\small
\begin{eqnarray*}
    g_1(A) &=& \log(1 + A) + A \log\left( \sqrt{\frac{A}{A + \frac{1}{9A} + 2}} \right) + \frac{1}{9A} \log\left(1 - \sqrt{\frac{A}{A + \frac{1}{9A} + 2}} \right)
     + \frac{1}{2} \log(2 \pi) - \frac{1}{2}\log(A + \frac{1}{9A} + 2) \\
    &=& \log(1 + A) + \frac{A}{2} \log(A) - \frac{1}{2}(1+A) \log(A + \frac{1}{9A} + 2)
    + \frac{1}{9A} \log\left(1 - \sqrt{\frac{A}{A + \frac{1}{9A} + 2}} \right)
    + \frac{1}{2} \log(2 \pi) \\
\end{eqnarray*}
}
Note that the function $p(A) = A + \frac{1}{9A}$ is increasing in $A$ for $A \ge \frac{1}{3}$.
We can conservatively bound $g$ from below noting $\frac{1}{9A} \le 1$ in our region.
{\small
\begin{eqnarray*}
    g_1(A) &\ge& = \log(1 + A) + \frac{A}{2} \log(A) - \frac{1}{2}(1+A) \log(A + 3)
    + \frac{1}{9A} \log\left(1 - \sqrt{\frac{A}{A + 2}} \right)
    + \frac{1}{2} \log(2 \pi) \\
    &\ge& \log(1 + A) + \frac{A}{2} \log(A) - \frac{1}{2}(1+A) \log(A + 3)
    - \frac{1}{9A} \sqrt{A} + \frac{1}{2} \log(2 \pi) =: \tilde{g}_1(A).
\end{eqnarray*}
}
We can use calculus to say that:
{\small
\begin{eqnarray*}
    \tilde{g}'_1(A) &=& \frac{1}{A+1} + \frac{1}{A+3} + \frac{\log(A)}{2}
    + \frac{1}{18 A^{3/2}} - \frac{1}{2}\log(A+3) \\
    &\ge& \frac{1}{A+1} + \frac{1}{A+3} + \frac{1}{18 A^{3/2}} + \frac{1}{2} \log(\frac{A}{A+3})
\end{eqnarray*}
}
This expression is monotone decreasing in $A$ and with a limit $\ge 0$.
Therefore $g_1(A) \ge \tilde{g}_1(A) \ge \tilde{g}_1(1/3)$ for all $A$.
We can explicitly evaluate this numerically and $\tilde{g}_1(1/3) > 0.01$ so we are done.
The final piece of this proof involves a similar argument for $g_2(A)$.
{\small
\medmuskip=1mu
\thinmuskip=1mu
\thickmuskip=1mu
\begin{eqnarray*}
    g_2(A) &=& \log(1 + A) + A \log\left( 1 - \sqrt{\frac{\frac{1}{9A}}{A + \frac{1}{9A} + 2}} \right) + \frac{1}{9A} \log\left(\sqrt{\frac{\frac{1}{9A}}{A + \frac{1}{9A} + 2}} \right) + \frac{1}{2} \log(2 \pi) - \frac{1}{2}\log(A + \frac{1}{9A} + 2) \\
    &=& \log(1+A) + A \log\left( 1 - \sqrt{\frac{1}{9A^2 + 18A + 1}} \right) + \frac{1}{2} \left(\frac{1}{9A}\log\left(\frac{1}{9A} \right) \right) - \frac{1}{2}\left(\frac{1}{9A} + 1\right)\log \left(A + \frac{1}{9A} + 2 \right) + \frac{1}{2}\log(2 \pi) \\
    &\ge& \log(1+A) + A \left(-\frac{1}{\sqrt{9A^2}}\right) + \frac{1}{2} \left(\frac{1}{9A}\log\left(\frac{1}{9A} \right) \right)
     - \frac{1}{2}\left(\frac{1}{3} + 1\right)\log \left(A + \frac{1}{3} + 2 \right) + \frac{1}{2}\log(2 \pi) \\
    &\ge& \log(1+A) - \frac{1}{3} - \frac{1}{2e} - \frac{2}{3}\log(A + \frac{7}{3}) + \frac{1}{2}\log(2 \pi) =: \tilde{g}_2(A)
\end{eqnarray*}
}
Once again we can see that $\tilde{g}_2$ is monotone increasing
\begin{eqnarray*}
    \tilde{g}'_2(A) = \frac{1}{1+A} - \frac{2/3}{A+7/3} = \frac{A + 5}{(A+1)(3A + 7)} \ge 0.
\end{eqnarray*}}
We complete the argument by noting $g_2(A) \ge \tilde{g}_2(A) \ge \tilde{g}_2(1/3) > 0.01$.
This concludes our proof of the PDF double crossing in region $R3$.
\qed

The results of Sections \ref{sec: r1}, \ref{sec: r2} and \ref{sec: r3} together prove Lemma \ref{lem: Gauss Beta}.
By proposition \ref{prop: ssd equiv}, Lemmas \ref{lem: Dir Beta} and \ref{lem: Gauss Beta} together complete the proof of Theorem \ref{thm: dir_norm}.

\section{Acknowledgements}
\label{sec: ack}

This work was generously supported by a research grant from Boeing, a Marketing Research Award from Adobe, and Stanford Graduate Fellowships, courtesy of PACCAR.

\newpage
{
\small
\bibliographystyle{plainnat}
\bibliography{reference}
}

\end{document}